\documentclass{Interspeech2024}
\usepackage{tabularx} 
\usepackage{arydshln}
\usepackage{multirow}
\usepackage{array}
\usepackage{hyperref}
\title{Sentence-wise Speech Summarization:\\Task, Datasets, and End-to-End Modeling with LM Knowledge Distillation}

\interspeechcameraready 

\name[affiliation={}]{Kohei}{Matsuura}
\name[affiliation={}]{Takanori}{Ashihara}
\name[affiliation={}]{Takafumi}{Moriya}
\name[affiliation={}]{Masato}{Mimura}
\name[affiliation={}]{\\Takatomo}{Kano}
\name[affiliation={}]{Atsunori}{Ogawa}
\name[affiliation={}]{Marc}{Delcroix}

\address{
  NTT Corporation, Japan
}

\email{kohei.matsuura@ntt.com~}


\begin{document}

\maketitle

\begin{abstract}
This paper introduces a novel approach called sentence-wise speech summarization (Sen-SSum), which generates text summaries from a spoken document in a sentence-by-sentence manner. Sen-SSum combines the real-time processing of automatic speech recognition (ASR) with the conciseness of speech summarization. To explore this approach, we present two datasets for Sen-SSum: Mega-SSum and CSJ-SSum. Using these datasets, our study evaluates two types of Transformer-based models: 1) cascade models that combine ASR and strong text summarization models, and 2) end-to-end (E2E) models that directly convert speech into a text summary. While E2E models are appealing to develop compute-efficient models, they perform worse than cascade models. Therefore, we propose knowledge distillation for E2E models using pseudo-summaries generated by the cascade models. Our experiments show that this proposed knowledge distillation effectively improves the performance of the E2E model on both datasets.
\end{abstract}

\section{Introduction}
Automatic speech recognition (ASR) has undergone significant advancements in the past decades \cite{prabhavalkar2024},
 primarily aiming to produce word-for-word transcriptions,
 but such transcriptions can be difficult to read for humans 
 due to spoken-style and redundant expressions. 
On the other hand, speech summarization (SSum) condenses a spoken document 
 into a concise and written-style summary 
 offering informative and easily digestible summaries, 
 and has gained increasing interest for processing speech from various domains, such as meetings \cite{rennard2023} and lectures \cite{kano2021}. 
However, unlike ASR, SSum is unsuitable for real-time applications 
 because it typically processes an entire spoken document at once,
 indicating the lack of technology to produce real-time and concise summaries of spoken content.

\begin{figure}[t]
  \centering
  \includegraphics[width=\linewidth]{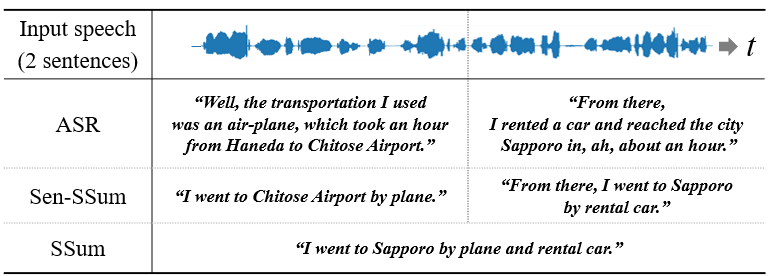}
  \caption{Examples of ASR, Sen-SSum, and SSum. Sen-SSum combines real-time processing and conciseness.}
  \vspace{-15pt}
  \label{fig:comp}
\end{figure}

To address this issue, we propose \textit{sentence-wise speech summarization} (Sen-SSum) to bridge the gap between ASR and SSum.
Figure \ref{fig:comp} compares ASR, SSum, and Sen-SSum with examples.
Sen-SSum goes beyond related technologies such as disfluency detection and removal \cite{lou2017} 
 and provides more concise and clearer outputs.
Furthermore, users can access summaries immediately,
 without waiting until the end of an entire meeting or lecture, 
 because Sen-SSum incrementally produces summaries after each speech sentence.
Therefore, as a substitute for personal notes, 
 Sen-SSum could help users review a meeting flow later 
 or catch up on the discussion when joining a lecture in the middle.
Despite its many promising applications, 
 the Sen-SSum task has not been well explored, partially due to the lack of publicly available datasets for Sen-SSum.

In this paper, we introduce a novel Sen-SSum dataset: \textit{Mega-SSum}.
The Mega-SSum dataset is based on the Gigaword dataset \cite{graff2003, rush2015} 
 and contains 3.8M English triplets of synthesized speech, transcriptions, and summaries.\footnote{\href{https://huggingface.co/datasets/komats/Mega-SSum}{\textrm{https://huggingface.co/datasets/komats/Mega-SSum}}}
We utilize a state-of-the-art multi-speaker text-to-speech model \cite{kim2021} to synthesize high-quality natural speech.
This dataset enables us to explore the impact of training data 
 availability on a large scale.
To increase the validity of our experiments with a more practical dataset, 
 we also use our internal Japanese Sen-SSum corpus, CSJ-SSum,
 which is based on the Corpus of Spontaneous Japanese \cite{maekawa2003}
 and contains 38k triplets with real speech.

We investigate two approaches for Sen-SSum: 1) cascade and 2) end-to-end (E2E) models.
The cascade model combines ASR and text summarization (TSum) models \cite{lewis2020}
 and can output high-quality summaries thanks to the TSum model pre-trained on large-scale text data.
The E2E model directly generates text summaries from input speech with a single encoder-decoder model \cite{sharma2022}.
This approach is promising in terms of parameter efficiency and potentially fast decoding,
 but it requires many expensive speech-summary pairs for training,
 and the lack of such large training dataset hampers the performance of E2E models.

To tackle this issue, 
 we propose knowledge distillation for the E2E models. 
We assume a practical scenario where a small subset of the training set (``core set'') 
 contains both summary and transcription labels,
 while the remaining samples lack either or both labels.
We increase the training data by creating pseudo-summaries from unlabeled speech
 using a cascade model trained on the core set. 
Subsequently, we train an E2E model using the core set and the pseudo-summaries.
We expect that the rich linguistic knowledge of the strong language model (LM), i.e., the TSum model, will be distilled into the E2E model via the pseudo-summaries.
The experimental evaluation showed that it significantly 
 improved the performance of E2E models on both datasets. 
Additionally, we found that the pseudo-summaries led to better summarization accuracies than the manual summaries in certain conditions.

\begin{table*}[t!]
  \centering
  \caption{Details of the Mega-SSum and CSJ-SSum datasets. Lang, dur, and CR denote language, duration, and compression rate, respectively. *The core set of Mega-SSum is included in the training set.}
  \vspace{-5pt}
  \begin{tabular}{c|cc|crrrrr} 
    \hline
    dataset                      & lang.               & orig. data & split      & \#samples  & \#speakers & total dur. (hrs) & ave. dur. (sec) & CR (\%) \\ 
    \hline
    \multirow{3}{*}{Mega-SSum} & \multirow{3}{*}{En} & \multirow{2}{*}{Gigaword} & train      & 3,800,000  &  2,559 &  11,678.2 &          11.1 &        26.2 \\
                              &                     &           & core set*  &    50,000  &  2,559 &     154.6 &          11.1 &   25.8\\
                              &                     & DUC2003   & eval       &       624  &     80 &       2.1 &          12.2 &    27.5 \\ 
    \hline
    \multirow{3}{*}{CSJ-SSum} & \multirow{3}{*}{Ja} & \multirow{2}{*}{CSJ} & train      &    38,515  &    726 &     115.8 &          10.8 &        43.1 \\ 
                              &                     &  & eval-CSJ &       467  &      9 &       1.4 &          10.8 &     42.8 \\ 
                              &                     & TED  & eval-TED &      1329  &      10 &      2.5 &             6.9 & 51.1\\
    \hline
  \end{tabular}
  \label{table:stats}
  \vspace{-15pt}
\end{table*}

\section{Related Work}
Text sentence summarization has been widely studied in the natural language processing (NLP) field \cite{rush2015,chopra2016}, but little research has dealt with the audio modality.
\cite{huang2022} is most closely related to our work, 
 where they experimented with Sen-SSum using a synthesized Gigaword dataset.
However, they investigated only cascade modeling and did not publish Sen-SSum datasets.
Furthermore, their dataset is ten times smaller than ours
 and has high word error rates (WERs), indicating poor speech quality.

The Sen-SSum task is also related to disfluency detection and removal \cite{lou2017} 
 since it excludes less important words and enhances the comprehensibility of ASR transcriptions. 
However, the primary goal of disfluency removal is not to make the document shorter. 
In fact, after eliminating disfluencies, the compression rates were 86\% for the Switchboard dataset \cite{godfrey1992} 
 and 92\% for the CSJ dataset, according to \cite{futami2023}.
Other related studies, such as spoken-to-written style conversion \cite{ihori2020, sunkara2021}, subtitling \cite{liu2020}, and other comprehensive post-processing \cite{neubig2012,liao2023}, have similar compression rates, though they further involve paraphrasing to generate more user-friendly transcriptions.
On the other hand, our Sen-SSum datasets introduced in Section \ref{sec:datasets} have lower compression rates of only 20\% (Mega-SSum) and 40\% (CSJ-SSum).
Thus, Sen-SSum is a significantly different task and more suitable for quickly grasping spoken content.

\section{Datasets for Sen-SSum}
\label{sec:datasets}
The availability of high-quality datasets is crucial for advancing
 research and development on the Sen-SSum task.
We introduce a novel dataset, \textit{Mega-SSum}, to investigate Sen-SSum. 
We also validate our experimental results on Mega-SSum using an in-house Japanese dataset, \textit{CSJ-SSum}.
Table \ref{table:stats} presents overviews and statistics of these datasets.

Mega-SSum is a large-scale English dataset for Sen-SSum, containing  3.8M synthesized spoken sentences and corresponding transcriptions and summaries.
It is based on the Gigaword dataset \cite{graff2003, rush2015},
 composed of the first sentences of news articles and their headlines,
 and is widely used for text sentence summarization studies \cite{chopra2016}.
The DUC2003 dataset \cite{over2007}, which is in a similar domain to Gigaword but with four manual summaries, rather than headlines, for each input sentence, is used for the evaluation set.
The summaries are fairly condensed, with a compression rate (CR)\footnote{The compression rate is defined as (\#words in summary) / (\#words in input text), 
 following \cite{goldstein1999}. Shorter summaries have lower rates.} of about 20\%.
To synthesize high-quality and natural-sounding speech from the text sentences,
 a multi-speaker text-to-speech model VITS \cite{kim2021} was trained from scratch using the LibriTTS-R dataset \cite{koizumi23},
 which is a sound quality improved version of LibriTTS \cite{zen19}.
The synthesized speech retains linguistic information, enabling us reasonable investigations for Sen-SSum.
For example, 
 when decoded with the Whisper (small.en) ASR model \cite{Radford2023} without fine-tuning,
 the WER on the evaluation set is only 7.8\%, which is as low as typical natural speech.
To simulate a realistic setting,
 the training set is split into a core set with the first 50k samples 
 and a remaining set with 3.75M samples.
The core set is used to investigate low-resource and practical situations.

We validate the results obtained on the Mega-SSum dataset using an 
 in-house Sen-SSum dataset named CSJ-SSum,
 which contains 38k real speech sentences, transcriptions, and summaries in Japanese.
It is based on the SPS subset of the CSJ \cite{maekawa2003},
 consisting of spontaneous monologues and corresponding transcriptions 
 on daily general topics such as commentary on recent news.
In addition, it has an out-of-domain evaluation set based on 10 Japanese TED talks (``eval-TED''). 
Professional annotators are employed to provide summaries following two instructions:
 1) Segment the speech and its transcription into sentences by manually inserting periods.
 2) Give a concise and written-style summary for each transcribed sentence, 
 retaining the most essential information.
This dataset enables us to confirm the possibility of Sen-SSum on real data 
 and for a different language. 
Incidentally, since this dataset is much smaller, 
 it justifies the practical need for our proposed knowledge distillation scheme in Section \ref{sec:prop}.

\section{Method}
\subsection{Cascade and end-to-end modeling}
We implement Sen-SSum with two distinct approaches: cascade \cite{lewis2020} and E2E \cite{sharma2022} speech-to-text summarization.

The cascade model combines the ASR and TSum models. 
The ASR model $\mathrm{ASR}(\cdot)$ first transcribes the input speech $\mathbf{x}$ 
 into its corresponding transcription,
and then the TSum model $\mathrm{TSum}(\cdot)$ predicts the target summary:
\begin{eqnarray}
    \hat{s} = \mathrm{TSum}(\mathrm{ASR}(\mathbf{x}))
\end{eqnarray}
 where $\hat{s}$ represents the summary hypothesis. 
We implement the ASR and TSum models using the Transformer-based encoder-decoder architecture \cite{vaswani2017}, trained with cross-entropy loss. 
The main advantage of cascade modeling lies in enabling the TSum model to acquire robust NLP capabilities by initializing it with a LM trained on a vast unlabeled text corpus.
Cascade modeling is a natural choice for Sen-SSum
 given that conventional SSum typically employs it \cite{murray2005,zhang2007_1}.

In E2E modeling, an encoder-decoder model $\mathrm{E2E}(\cdot)$ directly predicts
 the target summary from the input speech $\mathbf{x}$:
\begin{eqnarray}
    \hat{s} = \mathrm{E2E}(\mathbf{x})
\end{eqnarray}
Note that we build the E2E model by fine-tuning the ASR model, following \cite{sharma2022}.
E2E modeling is attractive with its compact and parameter-efficient structure,
 along with the potential for lower decoding latency.
Additionally, it mitigates the propagation of ASR errors
 and leverages the acoustic information in the input speech to predict summaries.
Despite its advantages, E2E modeling necessitates a substantial number of speech-summary pairs, which are expensive to collect and often limited, resulting in its poor summarization capabilities.

\subsection{Sequence-level knowledge distillation for E2E models}
\label{sec:prop}
To mitigate the training data scarcity for E2E models, 
 we propose leveraging a cascade model to generate additional training data.
Specifically, 
 the cascade model, trained on a small available dataset (i.e., the core set), generates pseudo-summaries 
 from unlabeled speech sentences 
 by transcribing them with the ASR model 
 and then summarizing the transcriptions with the TSum model. 
These pseudo-summaries, along with the human summaries in the core set, are used to train an E2E model. 
We anticipate that the rich linguistic knowledge embedded in the TSum model will be distilled to 
 the E2E model through the pseudo-summaries. 
Although this method is motivated by the success in the E2E speech 
 translation field \cite{gaido2020, inaguma2021} and widely known as \textit{sequence-level knowledge distillation} \cite{kim2016}, this is the first application to E2E SSum.


\subsection{Leveraging self-supervised models}
We also investigte the impact of integrating WavLM large\footnote{\textrm{https://huggingface.co/microsoft/wavlm-large}} \cite{Sanyuan2022}, a state-of-the-art 
 speech self-supervised learning (SSL) model 
 renowned for its effectiveness in low-resource spoken language understanding tasks, 
 on enhancing our E2E model with the Mega-SSum dataset.
Specifically, 
 we first use WavLM as a feature extractor,
 and then fine-tune it alongside the downstream E2E model.
Speech SSL models hold promise as potential alternatives to our proposed method, given their inherent linguistic knowledge from pre-training on vast speech-only data \cite{ashihara23}.

\section{Experiment}
\subsection{Setup}
\subsubsection{Model architectures}
We implemented three encoder-decoder models for ASR, TSum, and E2E Sen-SSum.
Both the ASR and E2E models comprised
 a 2-layer convolutional neural network with a sub-sampling rate of 4 for initial feature extraction,
 a 12-layer Conformer encoder with a model dimension of 512 and a kernel size of 31, 
 and a 12-layer Transformer decoder.
Both the encoder and decoder featured 2048-dimensional feed-forward (FF) layers.
Following \cite{matsuura2022}, 
 batch normalization layers in the Conformer blocks were replaced 
 with layer normalization layers,
 and learnable positional embedding was used for the decoder.
For the TSum model, 
 we used T5 models \cite{raffel2020}, consisting of a 12-layer Transformer encoder and decoder
 with a model dimension of 768 and 3072-dimensional FF layers.
The number of parameters in the cascade and E2E models were 362M (=142M+220M) and 139M, respectively.

\subsubsection{Training details: datasets and hyper-parameters}
For the experiments on Mega-SSum,
 we first trained an ASR model on the 960 hours of the Librispeech dataset \cite{panayotov2015} 
 and fine-tuned it using the 50k speech-transcription pairs in the core set.
We further fine-tuned the ASR model with the 50k speech-summary pairs to obtain an E2E model.
We denoted this baseline model as ``\textsf{E2E-base}''.
We prepared a TSum model by fine-tuning an English T5 model\footnote{\textrm{https://huggingface.co/google-t5/t5-base}} with 50k transcription-summary pairs. 
The combination of the ASR and TSum models were denoted as ``\textsf{Cascade-base}''.
To investigate the impact of training data volume, 
 we also prepared ASR, E2E, and TSum models
 trained with the first 100k, 500k, 1M, or 3.8M samples.
We denoted them as ``\textsf{E2E-HS}'' and ``\textsf{Cascade-HS}'' because they were trained with \textit{human summaries}, in contrast to ones trained with pseudo-summaries in Section \ref{sec:prop}.
The WERs of the ASR models were 11.7\%, 10.7\%, 9.2\%, 9.4\%, and 7.7\% for the evaluation set, respectively.

For the proposed knowledge distillation,
 we assumed that the core set with 50k samples  
 was fully available, and only the speech data of the first 50k, 450k, 950k, or 3.75M samples in the remaining set was additionally available.
The transcriptions and summaries were prepared as explained in Section \ref{sec:prop}.
The total number of training samples was 100k, 500k, 1M, or 3.8M since the 50k samples in the core set were added.
We denote these E2E models with the proposed knowledge distillation as  ``\textsf{E2E-KD}''.
To assess the effect of ASR errors on the proposed method,
 we also investigate the E2E models trained with pseudo-summaries generated from the reference transcriptions, 
 denoted as ``\textsf{E2E-KD (ref)}''.

For CSJ-SSum,
 we used the entire CSJ datasets \cite{maekawa2003} and 1M speech-transcription pairs from our in-house ASR datasets to train a ASR model.
The character error rates on the eval-CSJ/TED were 3.1\% and 16.7\%, respectively.
We then fine-tuned it with all 38k speech-summary pairs to obtain the baseline E2E model.
A Japanese T5 
 model\footnote{\textrm{https://huggingface.co/sonoisa/t5-base-japanese}} was fine-tuned with the 38k transcription-summary pairs to obtain the baseline TSum model.
For knowledge distillation, we leveraged the same 1M speech-transcription pairs used to train the ASR model.
We generated 1M pseudo-summaries from the reference transcriptions using the baseline TSum model,
 then used these to train the proposed E2E model alongside the 38k human summaries.
Note that the 1M pairs were selected from our in-house ASR dataset 
 to ensure that each reference transcription was longer than 10 characters and contained sentence-ending expressions, enabling the TSum model to produce reasonable summaries.

We adopted the same hyper-parameters for both Mega-SSum and CSJ-SSum unless specified otherwise.
To obtain an ASR model,
 we utilized the WarmupLR scheduler and the Adam optimizer, setting the maximum learning rate (LR) to 2x$10^{-3}$ 
 and the number of warmup steps to 40k.
The average batch size was set to 168 for LibriSpeech and 350 for CSJ.
We also applied the connectionist temporal classification (CTC) auxiliary loss with a weight of 0.3, 
 SpecAugment \cite{park2019}, and speed perturbation. 
The vocabulary size was set to 5,000 using byte-pair encoding \cite{sennrich2016} for Mega-SSum 
 and 3,262 with the character unit for CSJ-SSum.
For the fine-tuning, we adjusted the LR to 2x$10^{-4}$ and the number of warmup steps to 1k, reduced the batch size by one-fifth, and used the AdamW optimizer. 
CTC loss was omitted during E2E Sen-SSum model training.
To obtain a TSum model, we utilized a linearly decaying LR starting from 5x$10^{-5}$.
We used early stopping based on validation loss with a patience of 5 epochs. 

\subsubsection{Evaluation details}
In the evaluation, we consistently set the beam width to 4 for the ASR, E2E, and TSum models.
For objective metrics, we used ROUGE-L (R-L) \cite{lin2004} and BERTScore (BScr) \cite{zhang2020}, 
 which have been commonly used in previous TSum and SSum studies.
The R-L score measures superficial word matching, 
 while BScr leverages BERT embeddings to capture the semantic meanings of words.
Additionally, inspired by \cite{zheng2023}, we conducted A/B tests using the ChatGPT API\footnote{\textrm{https://platform.openai.com/docs/api-reference}} for a more comprehensive assessment.
Specifically, GPT4-turbo (gpt-4-1106-preview) was instructed to select the better summary of the two, considering the reference transcription.
 
\begin{table}[t!]
  \centering
  \caption{95\%-confidential intervals of ROUGE-L ($\uparrow$), BERT-Score ($\uparrow$), and CRs by cascade and E2E models on Mega-SSum.}
  \vspace{-5pt}
  \begin{tabular}{l|ccc} 
    \hline
    \multicolumn{1}{c|}{Model} & \multicolumn{1}{c}{ROUGE-L} & \multicolumn{1}{c}{BERTScore} & \multicolumn{1}{c}{CR (\%)}  \\ \hline
    Cascade-base  & 36.0\scriptsize{$\pm$1.5} & 62.6\scriptsize{$\pm$0.8} & 25.0\scriptsize{$\pm$0.7} \\ 
    E2E-base      & 30.7\scriptsize{$\pm$1.5} & 58.0\scriptsize{$\pm$0.8} & 21.3\scriptsize{$\pm$0.6} \\ 
    ~~+ WavLM       & 30.4\scriptsize{$\pm$1.5} & 58.2\scriptsize{$\pm$0.8} & 21.7\scriptsize{$\pm$0.6} \\ 
    E2E-KD$_{3.8\mathrm{M}}$   & 35.6\scriptsize{$\pm$1.6} & 61.9\scriptsize{$\pm$0.8} & 23.5\scriptsize{$\pm$0.6} \\      
    \hline
  \end{tabular}
  \vspace{-15pt}
  \label{table:result1}
\end{table}
\subsection{Results}
\subsubsection{Results on Mega-SSum}
\label{sec:results}
The first three columns in Table \ref{table:result1} show 
 the 95\% confidential intervals of R-L and BScr scores and CRs  
 by the cascade and E2E models trained on the core set.
As expected, the cascade model outperformed the E2E model
 due to the pre-trained TSum model's strong NLP capability.
In fact, the baseline E2E model occasionally generated only one or two words as a summary,
 likely due to its limited generalization capability, 
 resulted in lower CRs compared with the cascade model.
Although WavLM was also pre-trained on a large amount of unlabeled speech, 
 its combination with the E2E model did not improve the scores.
Its linguistic knowledge seemed insufficient to help the E2E model solve the summarization task.
The final column ``\textsf{E2E-KD}$_{3.8\mathrm{M}}$'' shows the performance of the E2E model trained with the 50k human summaries from the core set and an additional 3.75M pseudo-summaries. 
This proposed method significantly enhanced the E2E model to achieve a performance level comparable to that of the cascade model.

\begin{figure}[t]
  \centering
  \hspace{-1em}
  \includegraphics[width=1.00\linewidth]{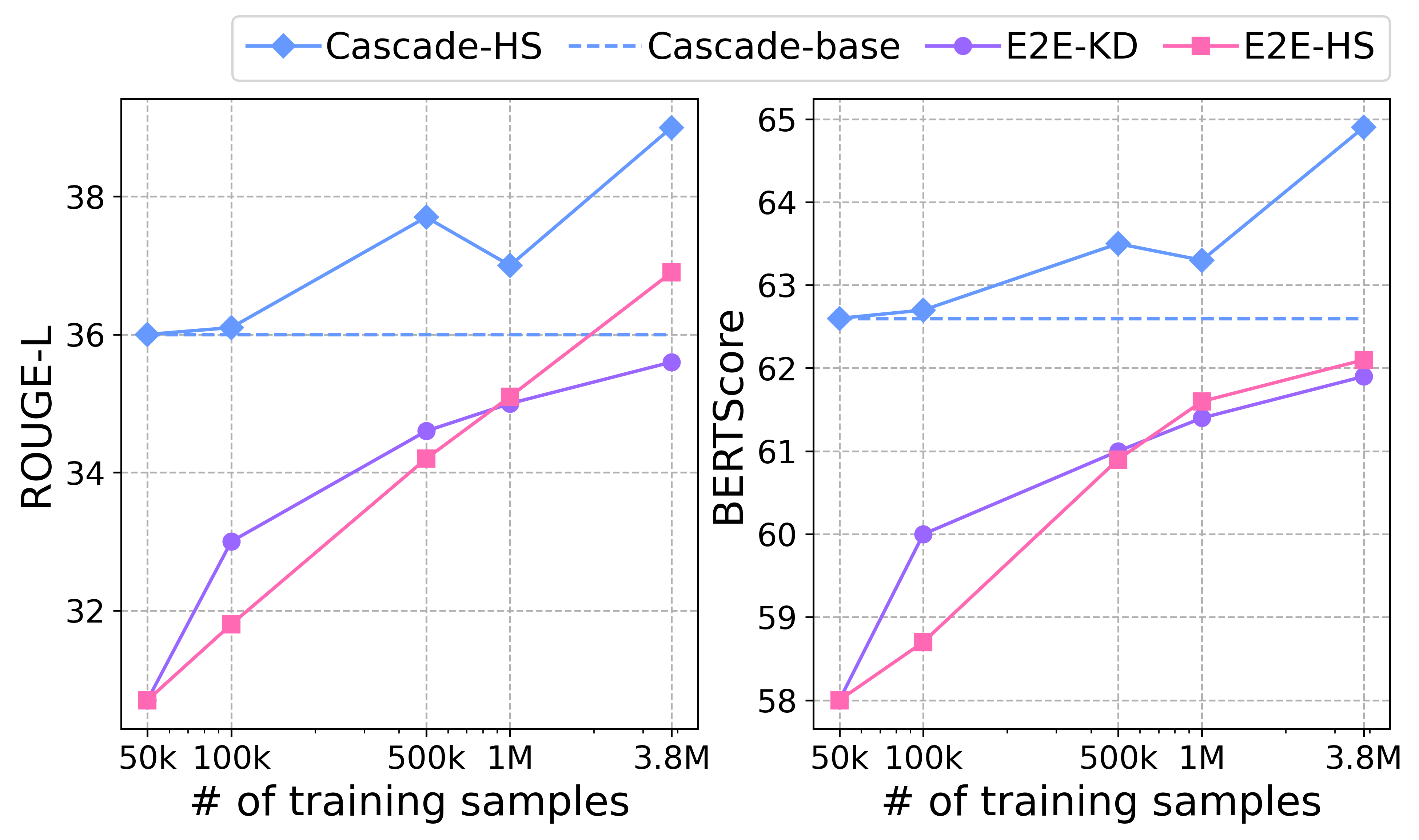}
  \vspace{-5pt}
  \caption{ROUGE-L ($\uparrow$) and BERTScore ($\uparrow$) by cascade and E2E models trained with various numbers of human summaries ``\textsf{*-HS}''. ``\textsf{E2E-KD}'' is the results of E2E model trained with pseudo-summaries (i.e., knowledge distillation).}
  \vspace{-5pt}
  \label{fig:result2}
\end{figure}

Figure \ref{fig:result2} illustrates the R-L and BScr scores
 when the cascade and E2E models were trained on varying amounts of training data,
 including the scores of the E2E models trained with pseudo-summaries.
As evident from the results of \textsf{Cascade-HS} and \textsf{E2E-HS},
 the cascade and E2E models generated better summaries with more training samples.
Additionally, the pseudo-summaries effectively improved the E2E model \textsf{E2E-KD}
 with its scores gradually approaching those of \textsf{cascade-base}, which generated the pseudo-summaries.

Interestingly, training with the pseudo-summaries resulted in better scores
 for the E2E model up to 500k samples.
This was likely because the pseudo-summaries were easier for the E2E model to learn compared with the human summaries.
For instance, the R-L score, which considers the longest common sub-sequence of two sequences, 
 between the pseudo-summaries and transcriptions (34.6$\pm$0.94, on the evaluation set)
 was significantly higher than that between the human summaries and transcriptions (27.2$\pm$1.03).
This suggests that the pseudo-summaries made Mega-SSum a more extractive summarization task, which is more similar to the ASR task and easier to reproduce.
However, with 1M or more human summaries, \textsf{E2E-HS} showed better scores than \textsf{E2E-KD} because it could reproduce more human-like summaries,
 which cannot be completely learned with erroneous and simplified pseudo-summaries.

\begin{figure}[t]
  \centering
  \vspace{-0pt}
  \includegraphics[width=0.95\linewidth]{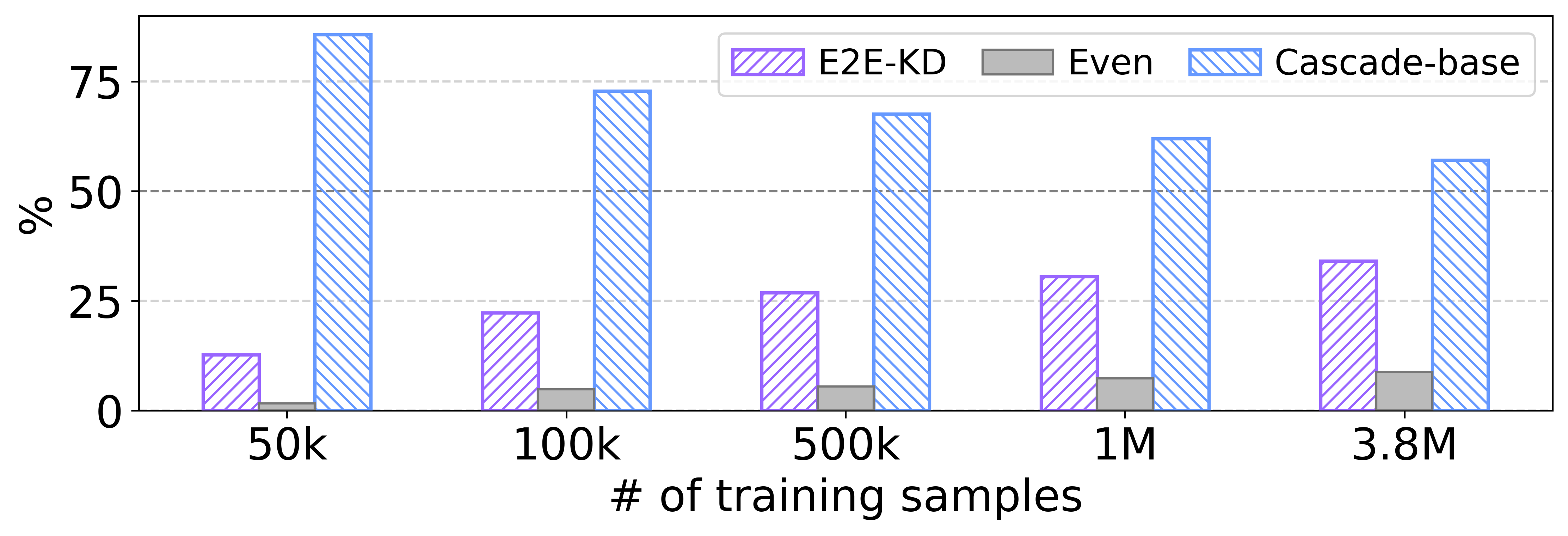}
  \vspace{-5pt}
  \caption{Results of the A/B test by ChatGPT. With more pseudo-summaries for training, more summaries by the E2E model were preferred to ones by the cascade baseline model.}
  \vspace{-15pt}
  \label{fig:result3}
\end{figure}

Figure \ref{fig:result3} shows the preference percentages of ChatGPT-conducted A/B tests
 comparing summaries generated by the baseline cascade model with those produced by the E2E models trained with varying amounts of pseudo-summaries.
As the dataset expanded, preferences gradually shifted towards the E2E model's summaries over those of the cascade model. 
However, even with 3.8M training samples,
 the E2E model's summaries were frequently deemed inferior to those of the cascade model.
This performance gap was not evident through conventional metrics such as R-L or BScr, indicating that more effective knowledge distillation methods may be required.

\begin{figure}[t]
  \centering
  \hspace{-1em}
  \includegraphics[width=1.0\linewidth]{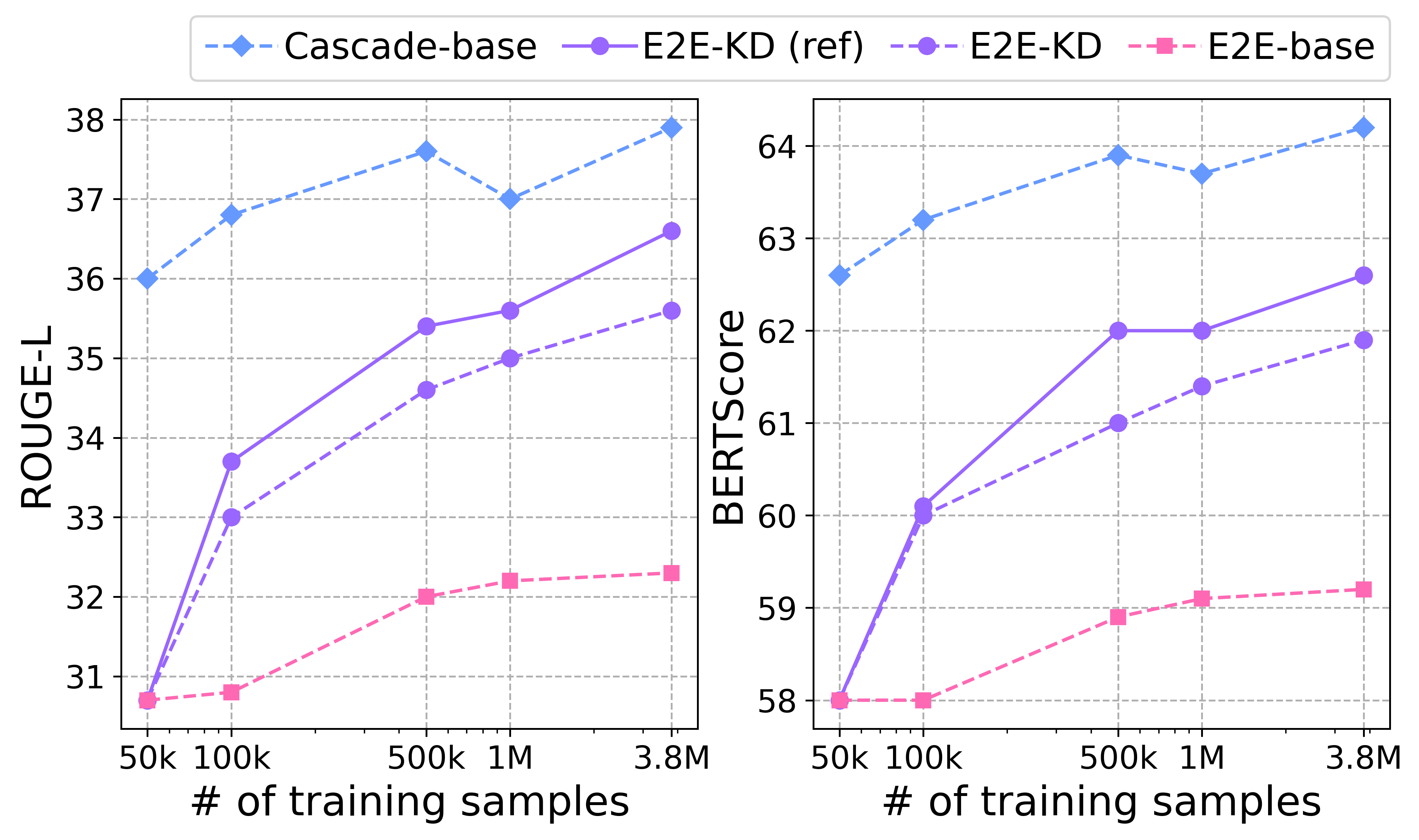}
  \vspace{-5pt}
  \caption{``\textsf{E2E-KD (ref)}'' shows the improvements by the knowledge distillation with reference transcriptions, which also improved the baseline models ``\textsf{*-base}''.}
  \vspace{-3pt}
  \label{fig:result4}
\end{figure}

Figure \ref{fig:result4} shows the effect of knowledge distillation
 when the reference transcriptions of the remaining set were also available in addition to the speech.
Note that the baseline cascade and E2E models, i.e., \textsf{Cascade-base} and \textsf{E2E-base}, were also improved 
 because the ASR model was trained on larger datasets.
While the speech-only data substantially improved,
 higher-quality pseudo-summaries derived from the reference transcriptions was important and resulted in better scores.

\subsubsection{Results on CSJ-SSum}
Table \ref{table:result5} shows the R-L and BScr scores on the CSJ-SSum dataset.
The scores were significantly higher than those for Mega-SSum
 because of more extractive nature and simpler structure of the summaries in the CSJ-SSum dataset.
Nevertheless, the proposed method significantly improved the scores on both evaluation sets,
 indicating its effectiveness in real-world scenarios.
We gained more improvements on the out-of-domain eval-TED set compared with the in-domain eval-CSJ set. 
This discrepancy may be attributed to the inherent generalizability of the proposed method, 
 which was better suited to adapt to diverse vocabularies in out-of-domain datasets. 

\begin{table}[t!]
  \centering
  \caption{95\%-confidential intervals of ROUGE-L (R-L) and BERTScore (BScr) by cascade and E2E models on CSJ-SSum. KD denotes our proposed knowledge distillation in Section \ref{sec:prop}.}
  \vspace{-5pt}
  \begin{tabular}{l|cc|cc} 
    \hline
    \multicolumn{1}{c|}{\multirow{2}{*}{Model}} & \multicolumn{2}{c}{eval-CSJ}  &  \multicolumn{2}{|c}{eval-TED}  \\ 
      & \multicolumn{1}{c}{R-L} & \multicolumn{1}{c}{BScr} &  \multicolumn{1}{|c}{R-L} & \multicolumn{1}{c}{BScr} \\ \hline
    Cascade       & 66.9\scriptsize{$\pm$2.1} & 84.7\scriptsize{$\pm$0.9} & 63.3\scriptsize{$\pm$1.2} & 82.6\scriptsize{$\pm$0.6} \\ 
    E2E           & 63.1\scriptsize{$\pm$2.3} & 82.8\scriptsize{$\pm$1.0} & 60.1\scriptsize{$\pm$1.3} & 80.7\scriptsize{$\pm$0.6}\\ 
    ~~+ KD & 65.7\scriptsize{$\pm$2.2} & 84.0\scriptsize{$\pm$1.0} & 63.1\scriptsize{$\pm$1.3} & 82.1\scriptsize{$\pm$0.6}\\ 
    \hline
  \end{tabular}
  \vspace{-14pt}
  \label{table:result5}
\end{table}

\section{Conclusion}
In this study, we introduced Sen-SSum along with two supporting datasets, Mega-SSum and CSJ-SSum. 
We demonstrated the potential of both cascade and E2E models 
 in Sen-SSum and the effectiveness of knowledge distillation 
 for E2E models using the cascade model. 
Future work could explore more efficient methods like \cite{matsuura2023},
 which directly integrates LMs into E2E SSum models and does not rely on pseudo-summaries.
It is also important to develop 
 context-aware models \cite{tiedemann2017}
 to consistently handle long speech documents in a sentence-by-sentence manner, which represents a more practical setting.

\newpage

\bibliographystyle{IEEEtran}
\bibliography{refs}

\end{document}